\documentclass{article}

\usepackage[utf8]{inputenc}

\usepackage{latexsym}
\usepackage{amsthm}
\usepackage{amsmath}
\usepackage{amsfonts}
\usepackage{amssymb}
\usepackage{tikz}
\usepackage{pgfplots}
\usetikzlibrary{shapes.multipart}
\usetikzlibrary{calc}
\pgfplotsset{compat=1.15}
\usepackage{subcaption}
\usepackage{float}
\usepackage[linesnumbered,boxed,commentsnumbered,ruled]{algorithm2e}
\usepackage{hyperref}
\usepackage{blindtext,titlefoot}

\numberwithin{equation}{section}

\theoremstyle{definition}

\title{A topological classifier to characterize brain states: \\ When shape matters more than variance}

\author{Aina Ferr\`a$\,{}^{1}$, 
Gloria Cecchini$\,{}^{1}$, 
Fritz-Pere Nobbe Fisas$\,{}^{1}$, \\[0.2cm]
Carles Casacuberta$\,{}^{1,\,2}$, 
Ignasi Cos$\,{}^{1,\,2,\,3}$}

\date{}

\begin{document}

\footnotetext[1]{Departament de Matem\`atiques i Inform\`atica, Universitat de Barcelona (UB), Gran Via de les Corts Catalanes, 585, 08007 Barcelona, Catalonia, Spain}

\footnotetext[2]{Institut de Matem\`atica de la Universitat de Barcelona (IMUB)}

\footnotetext[3]{Serra-H\'unter Fellow Programme, Catalonia, Spain}

\unmarkedfntext{This work was supported by the CERCA Programme of the Catalan Government (I.\,Cos); by the European Union’s Horizon 2020 Framework Programme for Research and Innovation under the Specific Grant Agreements No 945539, Human Brain Project SGA3 (I.\,Cos), and by MCIN/AEI/10.13039/501100011033 under grant PRE2020-
094372 (A.\,Ferrà) and projects PID2019-105093GB-I00 (I.\,Cos, A.\,Ferrà) and
PID2020-117971GB-C22 (C.\,Casacuberta).}

\unmarkedfntext{\emph{Keywords:} Topological data analysis; machine learning; classification; neuroscience; brain states; explainability}

\unmarkedfntext{\emph{Mathematics Subject Classification:} 55N31, 92C20, 62R40, 68T09}

\maketitle

\begin{abstract}
Despite the remarkable accuracies attained by machine learning classifiers to separate complex datasets in a supervised fashion, most of their operation falls short to provide an informed intuition about the structure of data, and, what is more important, about the phenomena being characterized by the given datasets.
By contrast, topological data analysis (TDA) is devoted to study the 
shape of data clouds by means of persistence descriptors and provides a quantitative characterization of specific topological features of the dataset under scrutiny.

In this article we introduce a novel TDA-based classifier that works on the principle of assessing quantifiable changes on topological metrics caused by the addition of new input to a subset of data. We used this classifier with a high-dimensional electro-encephalographic (EEG) dataset recorded from eleven participants during a decision-making experiment in which three motivational states were induced through
a manipulation of social pressure.
After processing a band-pass filtered version of EEG signals, we calculated silhouettes from persistence diagrams associated with each motivated state, and classified unlabeled signals according to their impact on each reference silhouette.
Our results show that in addition to providing accuracies 
within the range of those of a nearest neighbour classifier, the TDA classifier provides formal intuition of the structure of the dataset as well as an estimate of its intrinsic dimension. Towards this end, we incorporated dimensionality reduction methods to our procedure and found that the accuracy of our TDA classifier is generally not sensitive to explained variance but rather to shape, contrary to what happens with most machine learning classifiers.
\end{abstract}

\section*{Introduction}
\label{introduction}

The ability to capture detailed high-dimensional statistics of large datasets has earned deep learning the reputation of general solver for a wide range of complex problems
\cite{Grossberg2020,Sumathi2021,Yazdani2020}; 
from image analysis 
\cite{Moen2019,Wang2022}
to object recognition 
\cite{Pathak2018},
climate prediction 
\cite{Gibson2021,Rasp2018},
genomic analyses and behavioural prediction 
\cite{Zou2019}
or hidden variable identification for brain dynamics 
\cite{Yazdani2020}.
However, the downside of deep learning, as of any black-box AI technique, is the poor explainability of resuls and of the structure of the datasets, which necessarily bounds interpretability 
\cite{Ghassemi2021,Petch2021,Razavi2021}.
Possibly more puzzling, deep learning commits rare but unpredictable clustering errors, such as confusing a lion with a library, or a lamp with a traffic light 
\cite{Heaven2019}.
While odd, it is precisely the lack of a principled explanation leading to these unpredictable failures that makes of deep learning a technique with significant trust issues, in particular for AI in life-threatening decision-making scenarios.

By contrast, topology is a branch of mathematics devoted to characterize the structure of high-dimensional datasets by formal means 
\cite{LeCun2015,Nguyen2015}.
In brief, if a dataset can be represented as a 
point cloud in a hyperspace, topological data analysis (TDA) may characterize its connectivity, cycles within the cloud, or the number of clusters it contains, by means of a set of metrics (or summaries) that encompass all dimensions. 
Therefore, although not conceived for problem solving or to capture data variability per se, if we apply TDA to datasets of different classes within a classification problem, it will yield summaries specific to each class.
Consistent with this, some studies have proposed the use of TDA metrics as a preliminary stage to extract features for machine learning classifiers, e.g., for medical image analysis \cite{belchi,sizemore,lawson}, chemical components \cite{Bio1,Bio2}, or computer science problems \cite{Gebhart}. Thus, these classification problems do not operate in the multidimensional dataset or in a subsequent reduced dimensionality space \cite{PCA, ICA}, but rather in the domain of meaningful topological feature vectors \cite{exemple_tda1,exemple_tda2,exemple_tda3}, with a consequent interpretability gain.

While this is a promising research avenue that facilitates tracing class specificities back to the structure of data, our predicament is to use TDA as a classifier.
In brief, if different subsets of data (belonging to different classes) yield different topological summaries, it 
is plausible that the differences across summaries themselves 
can be directly exploited for classification purposes. Towards this end,  here we introduce a classifier inspired on this principle. 
The bonus of such a classifier with respect to a classical machine learning one is that, in addition to an accuracy and a confusion matrix, it should provide an informed intuition of the specific aspects of the dataset responsible for separability of classes.
Other approaches for classification or semi-supervised learning using topological persistence have been undertaken in \cite{Kindelan2021} and~\cite{RIOJA}.

As a testbed for our classifier, we used a challenging classification problem in dire need of explainability, which is the state cortical brain network during performance of specific tasks 
\cite{Allen2017,Cattai2018,Gilson2019}.
In particular, we focused on the characterization of the brain network of motivation, as defined in the context of a decision-making task between precision reaches 
\cite{Cos2021}
by different levels of social pressure. We analysed a set of electro-encephalograms (EEG) recorded as the decisions unfolded from eleven participants, by turning the problem into a three-class classification problem, in which we aimed at explaining the differences across these states on the grounds of our TDA classifier. We yield two main results: first, the TDA classifier obtained accuracies comparable to those obtained by a nearest neighbour classifier; second, accuracy strongly depended on the shape of the data but not on the amount of explained variance achieved by a dimension-reducing projection. In summary, specific topological descriptors indeed provide reliable ensemble characterizations of high-dimension neural states, and yields an avenue for data explainability complementary to machine learning.

\section{Materials and methods}
\label{section2}

\subsection{Persistence summaries}
\label{sec_persHom}
Topological data analysis is a branch of mathematics based on algebraic topology aiming to detect and represent structural features of datasets, such as sparseness, flares or cycles.
Its main tool is persistent homology~\cite{ELZ2002,ZC2005,Edelsbrunner2008,Carlsson2009}, an algebraic characteristic of simplicial complexes equipped with a real-valued filtering function. Persistent homology is well suited for describing the shape of a point cloud along a range of resolution scales.

Formally, a point cloud is a finite subset $X\subset\mathbb{R}^d$ for some $d\ge 2$ viewed as a metric space by means of the Euclidean distance. The \emph{Vietoris--Rips filtration} \cite{silva,ghrist} associated with $X$ is a nested family of abstract simplicial complexes $V_t(X)$ for $t\ge 0$, where $V_t(X)$ has a vertex for each point in $X$ and a $k$-simplex with $k\ge 1$ for each collection of points $v_0,\dots,v_k$ in $X$ such that $\|v_i-v_j\|\le 2t$ for all~$i,j$.

The inclusions $V_s(X)\subseteq V_t(X)$ for $s\le t$ induce  morphisms $H_n(V_s(X))\to H_n(V_t(X))$ for $n\ge 0$, where $H_n$ denotes $n$-dimensional simplicial homology~\cite{hatcher}. We compute homology with coefficients in a finite field ---we use the GUDHI Python Library \cite{gudhi} with a default choice of the field $\mathbb{F}_{11}$.
The \emph{birth parameter} $b$ of a homology generator in dimension $n$ is the smallest value of $t$ such that $H_n(V_t(X))$ contains the given generator, and the \emph{death parameter} $d$ is the smallest value of $t$ where that generator is mapped to zero. 
The \emph{persistence} or \emph{lifetime} of a homology generator is the difference $d-b$.

The \emph{persistence diagram} associated with the Vietoris--Rips filtration of $X$ in homological dimension~$n$ consists of all birth-death pairs $(b,d)$ for a basis of $n$-dimensional homology generators, drawn above the diagonal $y=x$ of the first quadrant in~$\mathbb{R}^2$. Points that are close to the diagonal (i.e., with a short lifetime) may correspond to inessential phenomena, while those with large lifetimes reflect persistent shape features of the given dataset. In some cases, however, the distribution of points near the diagonal carries relevant information that should not be neglected.

Persistence diagrams in dimension~$0$ contain information about connected components of Vietoris--Rips complexes, specifically about the way in which connected components merge as the parameter $t$ increases. Persistence diagrams in dimension~$1$ depict the appearance and disappearance of $1$-cycles, while persistence diagrams in dimensions $n\ge 2$ represent the evolution of $n$-dimensional cavities in the Vietoris--Rips complexes. 

Two fundamental 
results endow persistent homology with the robustness required for a rigorous mathematical theory with real-world applications.
The first one is the fact that persistence diagrams are well-defined \cite{ZC2005}, that is, 
do not depend on the choice of a basis of homology generators.
The second one is \textit{stability} \cite{cohen-steiner,CSO2014}, i.e., small perturbations in the data can only yield minor perturbations in the corresponding persistence diagrams. 

\begin{figure}[htb]
\centering
\includegraphics[width=\textwidth]{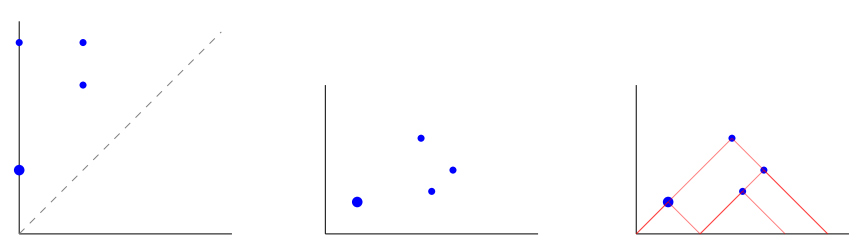}
\caption{Persistence landscape (right) obtained from a persistence diagram (left) by means of a $45^\circ$ rotation and rescaling (middle). Dots of larger size indicate multiplicity.}
\label{rotation}
\end{figure}

Dissimilarity between persistence diagrams can be measured by the \emph{bottleneck distance}, to which the stability theorem refers when claiming that two diagrams are close to each other.
The bottleneck distance between two diagrams $D_1$ and $D_2$ is defined as the largest coordinate-wise distance between points $x\in D_1$ and $\varphi(x)\in D_2$, where $\varphi$ is the matching that minimizes the outcome among all possible bijections between $D_1$ and~$D_2$, admitting points in the respective diagonals as eligible matches~\cite{ELZ2002}.
However, persistence diagrams equipped with the bottleneck distance are not well-suited for statistical analyses. On one hand, it is not feasible to compute averages of persistence diagrams
\cite{mileyko}; on the other hand, it is necessary for many purposes to work with representations of persistence diagrams in Hilbert spaces.
For this, a convenient summary of a persistence diagram is its \emph{landscape} \cite{bubenik,BUBENIK201791}.
A persistence landscape is a sequence of piecewise linear functions obtained by rotating the diagram $45$ degrees clockwise (Fig.\;\ref{rotation}) and choosing the $k$-highest point for each~$k\ge 1$ in the resulting figure \cite{bubenik}. 

More precisely,
for each point $(b,d)$ in a given persistence diagram, one considers a \emph{tent function}
$
\Lambda_{(b,d)}(t) = \max \{0, \min \{t-b, d-t\}\}
$
and defines
$\lambda_k \colon\mathbb{R}\to\mathbb{R}$ for each $k\ge 1$ as
$\lambda_k(t) = {\rm kmax} \{ \Lambda_{(b_i, d_i)}(t)\}$,
where $\{(b_i,d_i)\}$ is the set of all points in the given persistence diagram and kmax returns the $k$-th largest value of a given set of  numbers,
or zero if there is no $k$-th largest value. 
Consequently, $\lambda_k=0$ for sufficiently large values of~$k$.
The first landscape levels $\lambda_1,\lambda_2\dots$ represent the most significant features from the persistence diagram, while the last ones correspond to points near the diagonal and hence ephimerous phenomena.

\begin{figure}[htb]
\centering
\includegraphics[width=0.7\textwidth]{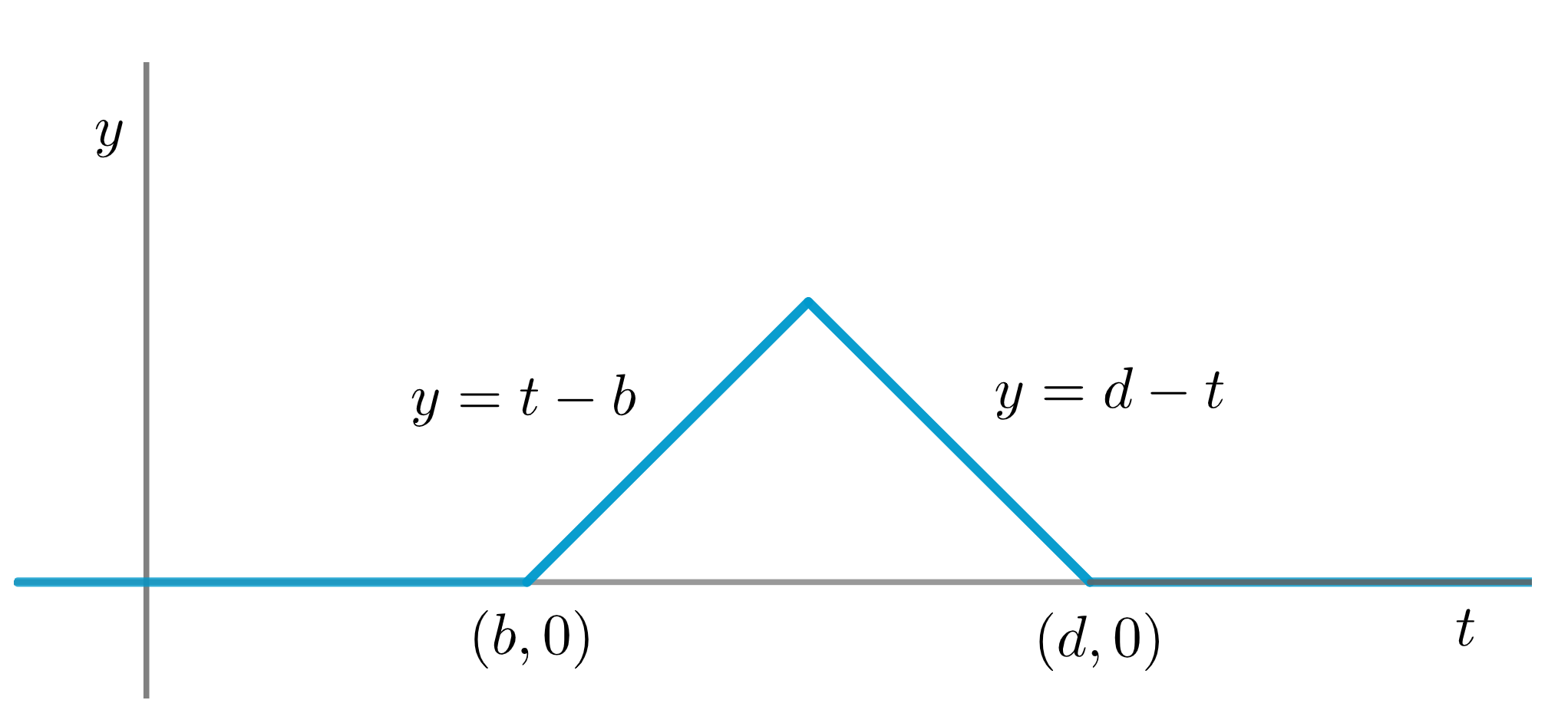} 
\caption{Tent function for a birth parameter $b$ and a death parameter $d$.}
    \label{landscapes}
\end{figure}

Despite their usefulness, persistence landscapes are memory-wise expensive when handling large datasets. 
To overcome this problem, \emph{silhouettes} were introduced in \cite{chazal} by considering a weighted average of the tent functions used to build a landscape:
\[
\phi_{w}(t)=
\frac{\sum_{i=1}^{m}\,w_i\;\Lambda_{(b_i,d_i)}(t)}{\sum_{i=1}^mw_i}.
\]
In this article we choose lifetimes $w_i=d_i-b_i$ as weights and use the resulting silhouettes as persistence summaries for our analyses.
Since lifetimes need to be finite, we discard points with infinite persistence in dimension zero.

\begin{figure}[htb]
    \centering
    \includegraphics[width=\textwidth]{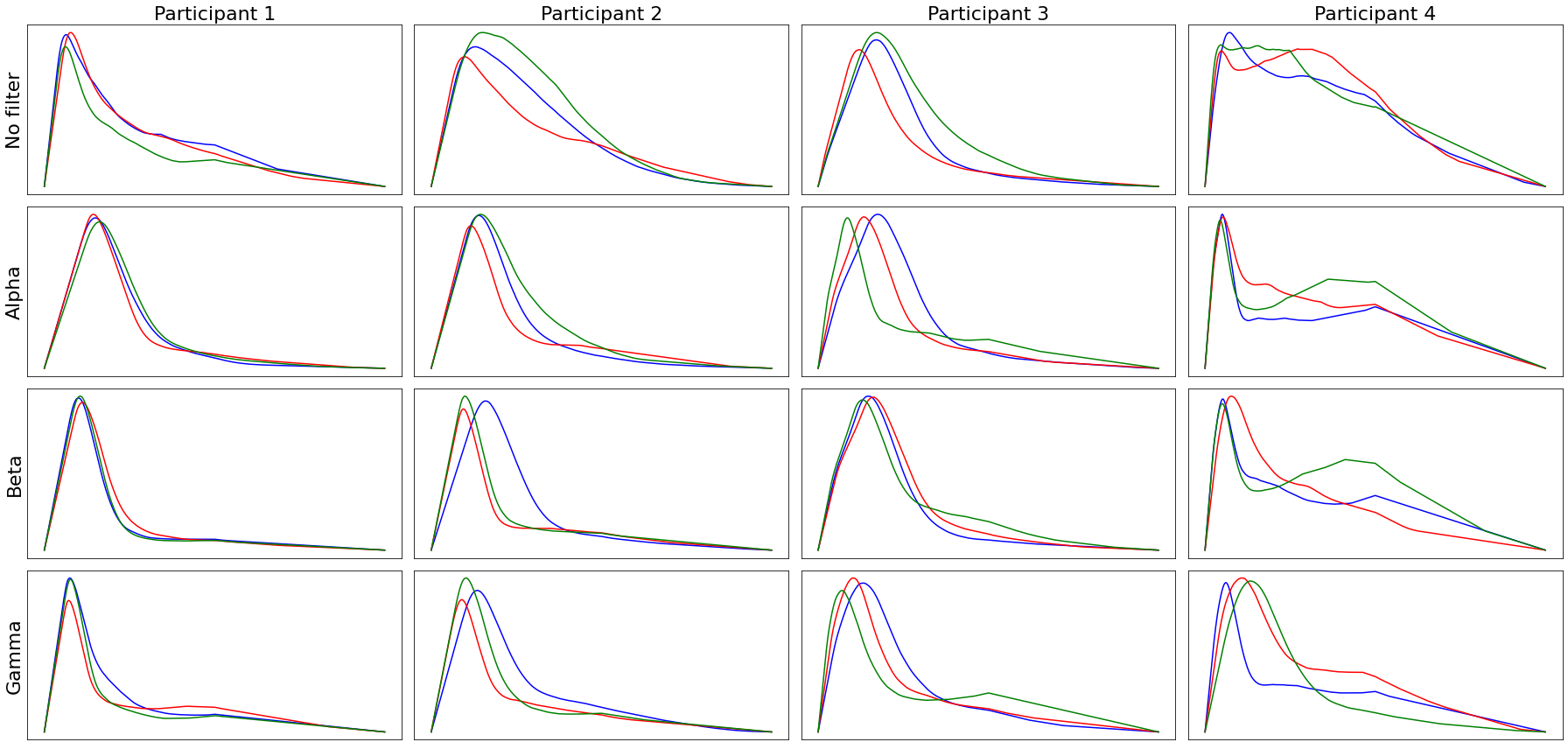}
    \vspace{0.1cm}
    \caption{Silhouettes from persistence diagrams in dimension zero for each motivational state ($M_0$: blue, $M_1$: red, $M_2$: green) for each frequency band (alpha, beta, gamma) plus the unfiltered dataset (no filter) for participants 1-4 in the space of sources without dimensionality reduction. The datasets are described in the next subsections.} 
    \label{fig:ex_sil}
\end{figure}

We measure dissimilarity between persistence diagrams by discretizing silhouettes into vectors of $1000$ components and computing Euclidean distances between such vectors.

Only persistence silhouettes corresponding to homological dimension zero were used for classification purposes. Higher dimensions were not considered in the study since generically no significant variation of persistence summaries can be expected by adding a single point to a training dataset, except in dimension zero.

\subsection{A topological classifier} 
\label{topological_clf}

A typical classification process starts with partitioning 
a labelled dataset $D$ into training and testing sets, 
assuming knowledge of the label
class $c$ to which each 
input $x$ belongs in the training set. Accuracy is defined as the percentage of correctly classified datapoints from the testing set.

Thus, if we have a dataset $D$ containing $m$ training datapoints as given by 
$D=\{(x_i,c_i)\}$
with $i=1,\dots,m$, the purpose of classifying is, for any given point $x$, to return its best predicted class~$c(x)$. 
Therefore, a critical question for a classifier algorithm is to define an appropriate metric of similarity between datapoints, so that similar points belong the same class.
Typically, given two points $x$ and $y$, similarity is quantified by a distance function $d(x,y)$ between them. For example, the \emph{nearest neighbour algorithm} uses the Euclidean distance.
Given a datapoint $x$ from the testing dataset, and given the training data 
$\{(x_i,c_i)\}$, the nearest neighbour algorithm
classifies $x$ as follows:
\begin{enumerate}
\item Calculate the Euclidean distance $d_i=d(x,x_i)$ of the point $x$ to each of the training points~$x_i$.
\item Find a point $x_{i_*}$ in the training dataset such that $d_{i_*}=\text{min}_i\,\{d_i\}$.
\item Assign the class label $c(x)=c_{i_*}$.
If there are equidistant points with different labels, the algorithm
selects the class containing the 
largest number of points. 
\end{enumerate}

Our persistence-based topological classifier follows instead the next procedure:

\begin{enumerate}
\item The training set is split into classes according to given labels.

\item For each class label $c$ in the training set, calculate the corresponding silhouette $S_c$ in homological dimension $0$ with lifetimes as weights.

\item To classify an
input $x$ from the testing set, add $x$ to the cloud of training datapoints $X_c$ of each class label~$c$.
Then, recompute the persistence silhouettes $S_{c,x}$ for the datasets $X_c\cup\{x\}$,
and finally calculate the 
Euclidean distance $d(S_c,S_{c,x})$ between the newly obtained silhouettes and the former ones.
\item
Assign the class label $c(x)=c_*$ 
where $c_*=\text{argmin}_c\,\{d(S_c,S_{c,x})\}$.
\end{enumerate}

The underlying assumption is that point clouds from different classes exhibit recognizably different shapes. More precisely, if we add a point to the point cloud of the correct class, 
the resulting topology should not fundamentally change, meaning that all distances should remain small. By contrast, if a point is added to a point cloud of a different class, then the topology should be altered more visibly.

\subsection{Time series of brain states of motivation}

For practical purposes, we opted for testing our TDA classifier on a dataset previously recorded,
rather than opting for recording a new one ad-hoc to this end. The dataset and the context of its recording
is described next. In addition to having an advantageous set of baseline results to fall back on for comparison purposes, which would undoubtedly facilitate the validation of the TDA classifier, we specifically chose this 
dataset to task the TDA classifier to provide a richer characterization of the dataset underlying structure 
than machine learning algorithms, and of the processes generating that dataset.

Reward and motivation are two fundamental drives of human behaviour. Consistent with this, a large number of studies in neuroscience have intended a careful identification and characterization of the brain centers of reward 
processing, most often based on analyses of functional magnetic resonance imaging (fMRI) recordings in humans.
However, one of the questions that remains to be fully answered is how the different expressions of motivation 
are distributed across the brain network. In other words, what is the distribution of the brain network of social motivation? 

To answer this question, we refer to a study carried out with eleven participants in Barcelona 
\cite{Cos2021},
in which high-density electro-encephalograms (EEG) were recorded from human participants during a decision-making task
in which motivation was modulated via social pressure. 
The goals of this task were the formal characterization of the influence of social pressure on movement 
decisions and on choices of movement parameters to make precision reaches and of the brain network of social motivation 
\cite{Cos2021}.
The \emph{manipulation of motivation} was performed by means of social pressure, as a function of the 
participant’s aiming accuracy with respect to that of a virtual partner. In brief, the idea was that
the presence of the partner would introduce an implicit bias to boost the participant to improve their accuracy. We informed the participant that they
would perform within a community of players, 
and that at each block, 
they would have a different partner from this community to perform alongside. 
To reinforce the belief on the existence of a partner, at the end of each trial we showed a horizontal 
green bar displaying the accuracy attained, normalized between 0 and 100\%, over a red bar displaying 
the partner's accuracy. The purpose of the simulated partner was to introduce an implicit bias to modulate 
the participant’s motivation to reach more accurately. Thus, we instructed each participant not to compete 
to just focus on 
their own performance, and to disregarding the partner. To parameterize the bias introduced 
by the presence of the virtual partner, we used partners of two types, less or more accurate that the participant (Fig.\;1D). We also matched the gender of partner to that of the participant to control for cross-gender 
effects.

The level of social pressure remained constant throughout each block, maintaining the same virtual partner.
The partner was varied across each block in a counter-balanced fashion. Each participant performed 
two sessions of six blocks each, with each block consisting of one-hundred and eight trials. 
The six blocks were distributed into two groups of three. Each group consisted of one block 
solo and two blocks each alongside a partner of a lesser/higher aiming skill. 
The goal of this manipulation was to induce three distinct motivated states as a function of 
the level of social pressure 
exerted:
\begin{itemize}
\item Solo, the participant performed alone. No social pressure in this condition.
\item Easy, the participant performed alongside a virtual partner of a lesser skill than
them.
Some social pressure was expected in this condition.
\item Hard, the participant performed alongside a virtual partner of a higher skill than 
them.
High pressure was expected in this condition.
\end{itemize}

\begin{figure}[ht!]
    \centering
    \includegraphics[width=0.6\textwidth]{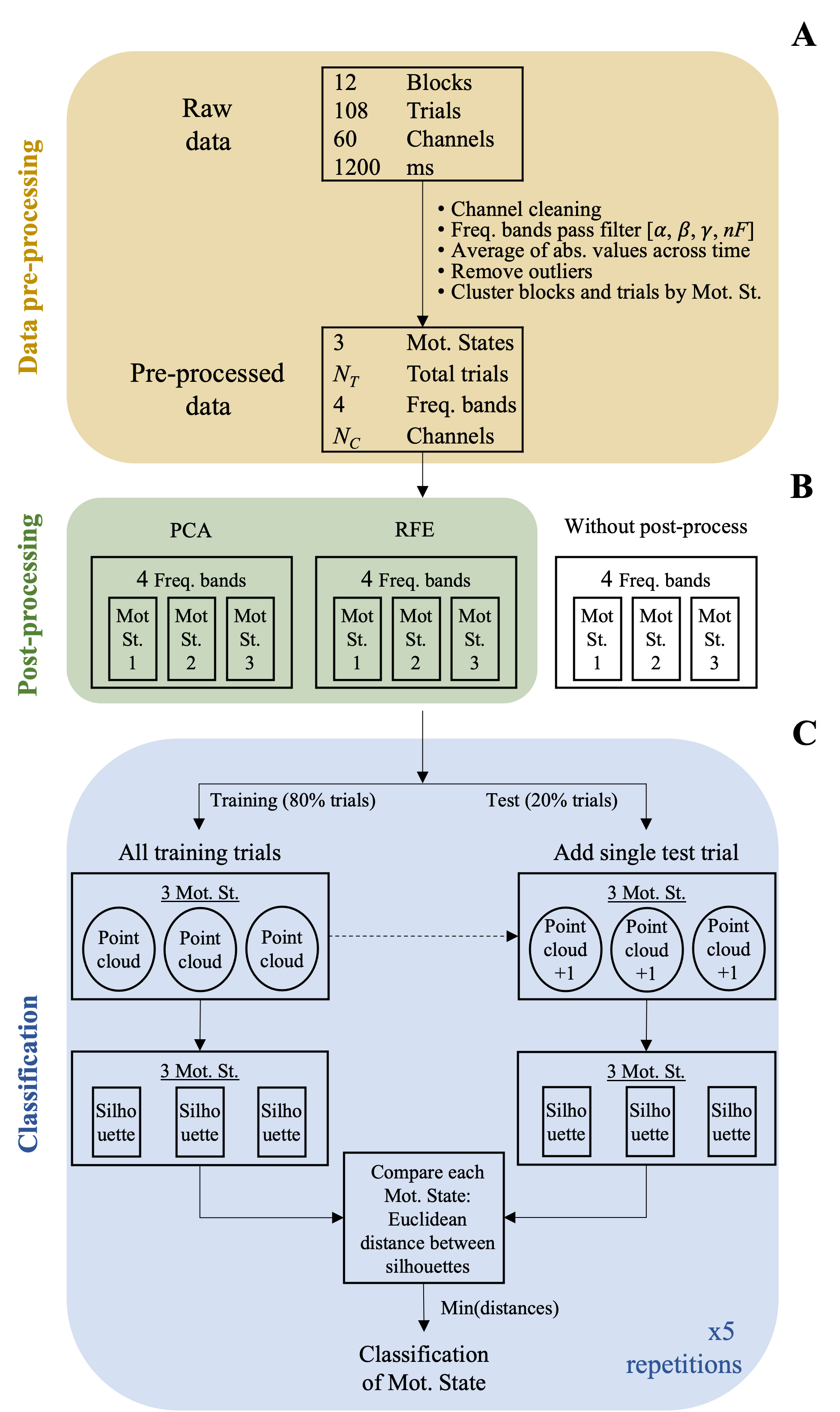}\vspace{0.2cm}
    \caption{\textbf{A. Pre-processing schematic.} Channels are either electrodes or sources.
    Each channel of raw data is band-passed filtered into the three typical EEG bands ($\alpha$: low; $\beta$: medium; $\gamma$: high). Channels containing artifacts or noisy information are removed, resulting in a final number of $N_C$ channels. Absolute values of time series are averaged for each channel. Outliers are removed from the resulting dataset. \textbf{B.~Post-processing schematic.} Principal component analysis (PCA) or recursive feature elimination (RFE) are used. No post-processing is performed on raw data. \textbf{C.~Classification schematic.}  80\% of the data is used for training purposes and 20\%
    for testing ones. 
    Persistence silhouettes are used for classification.}
    \label{fig:pipe2}
\end{figure}

The dataset used for TDA analysis recorded during performance of this task. 
The \emph{structure of the dataset} consisted of EEG fragments from each of the participant's recordings, 
for each level of induced social pressure. Since we were specifically interested in assessing the changes
on the motivation baseline, we selected an interval of interest of 1200~ms at each trial, starting 800~ms 
before the first stimulus onset ---the initial cue, and ending 400~ms later (see Fig.\;\ref{fig:pipe2}A)--- before any 
movement or stimuli were presented on the screen at that trial.
Each channel was filtered with a 4th order notch filter at 50, 100 and 150~Hz to remove electrical interference 
caused by the power supply line, and a 4th order bandpass Butterworth filter between 0.1 and 100~Hz to constrain 
the band of interest.
Electrodes with EEG level exceeding either 200~V or voltage step/sampling 50~V within intervals of 200~ms 
were removed from further analysis. Baseline was corrected, and the datasets were $z$-scored by using the 
recordings during trial block types 1 and 2 of each session ---when the participant was playing solo, as 
a reference for baseline motivation. Eye related artefacts were removed by means of independent component 
analysis (ICA), implemented in with custom-made Field-trip open-source toolbox (www.fieldtrip.com) and EEGLAB 
scripts (http://sccn.ucsd.edu/eeglab, UC San Diego, CA, USA). 
The procedure to identify eye-movement related sources was semi-automatized,
first correlating each source obtained with the signal from the electrodes recording eye movements to obtain 
a first metric of relatedness. Second, we visually inspected all sources to corroborate that their shape and 
spatial location matched those of ocular artefacts. Eye-related sources were removed and the cleaned signal 
obtained by inverting the ICA process. A final source space was obtained by applying ICA again, but forcing 
the resulting space to be of smaller dimension than the dimension of the electrode space, in order to capture 
a few independent areas of the brain whose signals are sent to the electrodes. Thus the dimension of the source 
space may be different for each participant ---usually less than~50. 

The computation of the projection of the data on the source space was performed with custom-made MATLAB scripts 
based on the EEGLAB library, combined with the electrode spatial location map. The Brain Products Unicap 64 
(Brain Products GmbH, Gilching, Germany) configuration was used to establish a spatial reference between 
the electrode placement and to perform source localization. We assumed a spherical head model. 
Please, note that ICA projections serve a dual purpose: first, to distribute the information contained in the 
electrode signals along directions of maximal inter-independence between dimensions. Second, to provide a rough
anatomical estimate for the location of the brain sources generating the signals recorded by the electrodes, on 
the brain cortical surface.

In summary, the dataset consists, per trial, of a variable number of channels (electrodes or sources), denoted as 
$N_C$, lasting $1200$~ms each. Each participant performed $12$ blocks of $108$ trials. Note that $12$ blocks were 
recorded during two different sessions, containing $6$ blocks each, balancing out the number of motivated blocks.

\subsection{Analysis pipeline}
\label{sec_pipeline}

For each of the eleven participants in the study
(four male and seven female aged $55\pm 5.8$)
the dataset consisted of 
$1200$ ms $\times$ $60$ electrodes EEG segments, repeated over $12$ blocks (six per session) of $108$ trials 
each (Fig.\;\ref{fig:pipe2}A). The level of social pressure leading to a specific motivated state was maintained 
constant across each block, and there were four blocks for each motivated state (two each session).

\subsubsection*{Pre-processing}
\begin{itemize}
    \item First, data were visually inspected to identify and remove noisy or artefactual channels from further analysis. We call $N_C$ the number of channels surviving this process.

\item Second, EEG signals encompass a spectrum ranging $0.01$--$100$ Hz, which we distributed 
into the three typical frequency bands:
$\alpha$ ($8$--$15$ Hz), $\beta$ ($15$--$32$ Hz), $\gamma$ ($32$--$80$ Hz), obtaining three
band-passed versions of the original temporal series. Each of these signals were 
analyzed in the same fashion, and independently of each other, alongside with an 
unfiltered baseline version of the original signal.
This distribution of frequency bands responds to the established association 
of brain function with power fluctuations in specific bands and electrode locations.
Band-pass filtering was performed with custom-made scripts in Python, using the 
\verb|iirfilter()| function from the \verb|scipy.signal| library \cite{scipy}.
Previous analyses on previous studies suggest that motivation related modulations
belong in the high-gamma band, thus suggesting that motivation related biases 
should be better encoded in the higher frequency band 
\cite{Cos2021}.

\item Third, we averaged each electrode or source absolute value across the 1200~ms window of 
observation, obtaining a dataset organised as a matrix of $N_T$ trials by $N_C$ channels.
Outliers exceeding twice the standard deviation from the average of norms of datapoints
were disregarded from further analysis.

\end{itemize}

Our classification was conducted within the signals obtained for each band-pass independently, 
as well as with an unfiltered version of the original signal.

\subsubsection*{Post-processing}
Two different dimensionality reduction methods were used to the pre-processed dataset by means of the 
\verb|sklearn| library \cite{sklearn}, yielding dimensions between $2$ and $10$.

\begin{itemize}
\item Principal component analysis (PCA) is a linear projection onto a lower-dimensional space of principal 
components, where the first principal component of a point cloud is the one that explains the most variance, 
and each successive principal component explains the most variance in what is left once the effect of the 
previous components is removed \cite{PCA}.

\item Recursive feature elimination (RFE) consists of successively removing coordinates with the lowest impact on the 
accuracy of a classifier \cite{RFE}. In our study, RFE was applied to the pre-processed dataset using a logistic 
regression model to assign weights to the features.
\end{itemize}

\subsubsection*{Classification}
Once each dataset (for each frequency band) was properly formatted, it was input into the TDA classifier 
(Fig.\;\ref{fig:pipe2}B).
From this point on, trials were considered as points of a cloud to be classified. 
The classification operation 
was carried out as described in the topological classifier subsection.
Each classification was repeated $5$ times and the resulting accuracies averaged over the $5$ repetitions. 
At each repetition, post-processed datasets were partitioned into 80\% training data and 20\% testing data. 
In the case of raw data (no dimensionality reduction), pre-processed datasets were used.

The analyses were performed both with the original signals in electrode space, as well as with those projected
onto the brain source space.

\section{Results}
\label{section3}

To first establish a baseline of accuracy that would enable an assessment of the influence of 
dimensionality reduction techniques on the classificacion process, we tested our 
TDA classifier on the original EEG signals projected on source space; 
see Analysis pipeline.
Second, we tested the influence of two dimensionality reduction methods (PCA and RFE) on the
classification process.
Third, to further assess the influence of the ICA projection on the classifier, we 
also tested the original dataset, as collected in electrode space. 
Each of the aforementioned tests were performed with each of the band-passed version of the EEG signals
($\alpha$, $\beta$, $\gamma$), independently; see Analysis pipeline. 

\begin{figure}[htb]
\centering
\begin{subfigure}{\textwidth}
\centering
\includegraphics[width=0.47\textwidth]{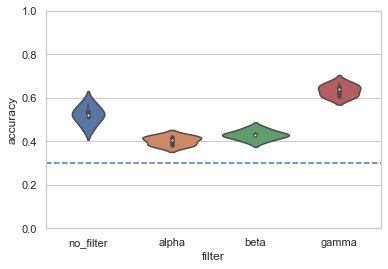} \hspace{0.2cm}
\includegraphics[width=0.47\textwidth]{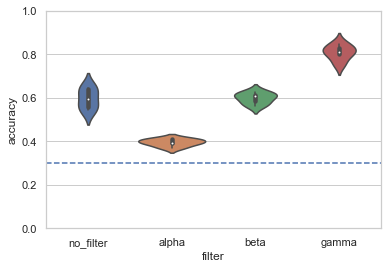}
\end{subfigure}
\begin{subfigure}{\textwidth}
\centering
\includegraphics[width=0.47\textwidth]{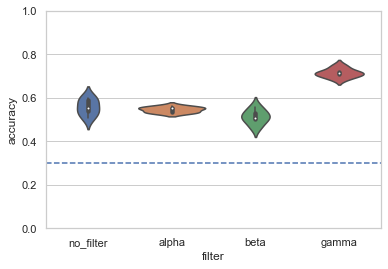} \hspace{0.2cm}
\includegraphics[width=0.47\textwidth]{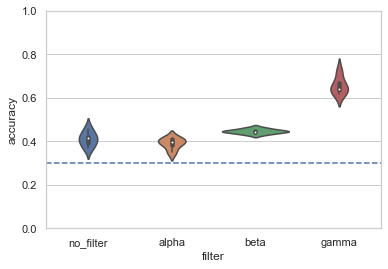}
\end{subfigure}
\vspace{0.2cm}
\caption{Accuracies of the topological classifier on source space by frequency band using the space of sources and without dimensionality reduction for participants 1, 3 (top right), 7 and 11. A complete set of results for all participants is given in the Appendix.}
    \label{Fig3-1}
\end{figure}

\subsubsection*{Classification on source space}
We performed the classification for the dataset of each participant within each frequency band, using the data projected
onto source space prior to any dimensionality reduction. Fig.\;\ref{Fig3-1} shows the classification accuracies obtained 
by the TDA classifier for four typical participants. 
Although the classification yielded some differences across participants, the main result obtained is a consistent 
top accuracy in the gamma band for all participants but one, ranging in average between 60\% to 80\%.
Violin plots in Fig.\;\ref{Fig3-1} encode median, interquartile range, and a kernel-smoothed probability density of the data.
The violin plots were performed using the \verb|seaborn| library \cite{seaborn}.

\begin{figure}[htb]
\centering
\includegraphics[width=\textwidth]{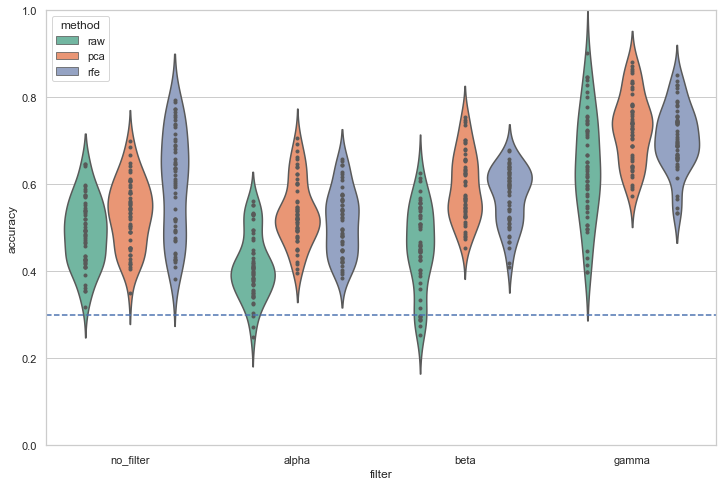}
\vspace{0.1cm}
\caption{Comparison of baseline accuracies (\textit{raw})
of the topological classifier on source space for each frequency band with accuracies obtained after dimensionality reduction with principal component analysis (\textit{pca}) and recursive feature elimination (\textit{rfe}), for participants 1 to 11. Each violin contains accuracy percentages of five repetitions for each participant.}
\end{figure}

Using this as a baseline, we assessed the influence of dimensionality reduction on the classification. 
The underlying hypothesis is that, if the TDA classifier focuses on elements of the shape of the point cloud to determine separability, datasets rearranged in specific spaces should yield point clouds with shapes more favourable to be classified. To test this, we performed two complementary dimensionality reduction techniques.
For each participant dataset, we first performed a principal component analysis (PCA) decomposition, selecting the dimensions that would explain until 95\% of data variance for all participants, ultimately retaining from 2 to 10.

In a complementary fashion, we performed a second dimensionality reduction technique on the same datasets.
Unlike PCA, recursive feature elimination (RFE) is based on assessing the contribution of specific components of the original dataset to the classification process. When performed in source space, this results in a ranking of sources. The datasets resulting from PCA and RFE were also classified for each participant and for each frequency band.
The summary results obtained from these classifications are listed in Fig.\;\ref{Table3-2}, 
showing the accuracies of the topological classifier for each of the eleven participants, for each frequency band and for each dimensionality reduction method. The accuracies in 
Fig.\;\ref{Table3-2} have been obtained by averaging the accuracies obtained from five repetitions for each dimension. Since there are three distinct classes, chance level equals~$0.33$.

These analyses yielded two main results. First, consistently with the baseline classification results of the TDA classifier and with the results of previous machine learning classifiers, the highest accuracies 
were obtained in the gamma band for all participants but one 
\cite{Cos2021}.
Second, consistently with the hypothesis of dimensionality reduction rearranging the point cloud in shapes that facilitate classification, our results consistently confirm that point clouds in reduced dimensionality spaces yield higher accuracies than the baseline, high-dimensional dataset.

\subsubsection*{Effect of dimensionality reduction}
Since higher classification accuracies are obtained in low dimensional spaces, a question on how 
accuracy relates to dimension arises. In the case of RFE, we aim to find an optimum number of sources as guided by previous results
\cite{Cos2021}.
In order to do so, we tested our topological classifier using the spaces resulting of applying recursive feature elimination to the EEG data projected onto the space of sources, increasing the number of selected sources by the algorithm from 2 to 10. Classifications were performed for each participant and each frequency band in parallel.

The case of PCA presents a more involved study of the effect of the dimensionality mainly due to the relationship between accuracy and explained variance. In most machine learning classifiers, there is a direct relationship between the two: the higher the explained variance, the higher the accuracy, often converging once $95\%$ of explained variance is reached. Our hypothesis is that the notion of shape that TDA techniques are able to capture from the data is independent of its variability. Instead, we expect to find an optimum number of principal components such that the resulting projected spaces yield shapes more favourable to be classified. We tested our topological classifier using the spaces resulting 
from principal component analysis applied to the space of sources, increasing the number of selected principal components at each step from 2 to~10. We recorded the accuracy of each classification as well as the amount of explained variance by the number of components at that step. Classifications were performed for each participant and for each frequency band in parallel. To further assess the effect of independent component analysis to the shape of the data, we also performed the same tests using the space of electrodes, namely the EEG data without applying the ICA algorithm.

\begin{figure}[htb]
\centering
\begin{subfigure}{\textwidth}
\centering
\includegraphics[width=0.47\textwidth]{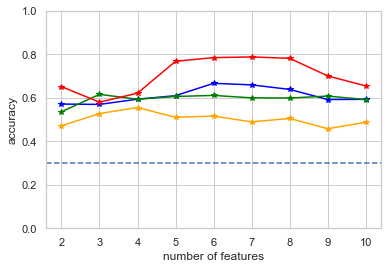} \hspace{0.2cm}
\includegraphics[width=0.47\textwidth]{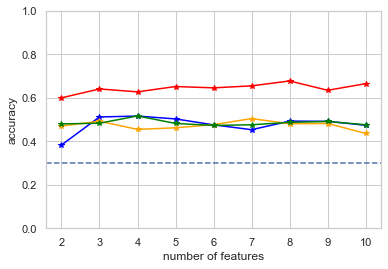}
\end{subfigure}
\vspace{0.1cm}
\caption{Variation of accuracy of the topological classifier by frequency bands (blue: no filter; yellow: alpha; green: beta; red: gamma) depending on the number of sources selected by recursive feature selection for participants 1 and 2. The blue dotted line indicates chance level.}
    \label{Fig3-4}
\end{figure}

Figure \;\ref{Fig3-4} shows the variation of accuracy of two typical participants with respect to the number of sources selected by RFE (from 2 to 10), separating frequency bands. Higher accuracies are consistently found within the gamma band. The accuracy evolution graph shows that, in general, only a low amount of sources is needed to obtain the best accuracy. More precisely, five sources are enough for the classification task for all participants, which is consistent with the results obtained in \cite{Cos2021}.

\begin{figure}[htb]
\centering
\begin{subfigure}{\textwidth}
\centering
\includegraphics[width=0.47\textwidth]{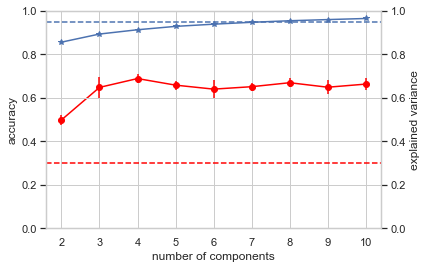} \hspace{0.2cm}
\includegraphics[width=0.47\textwidth]{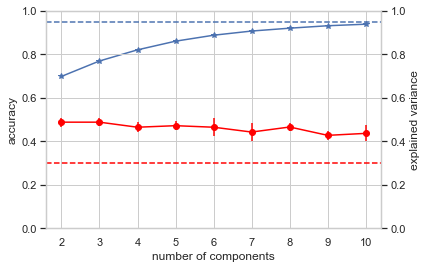}
\end{subfigure}
\begin{subfigure}{\textwidth}
\centering
\includegraphics[width=0.47\textwidth]{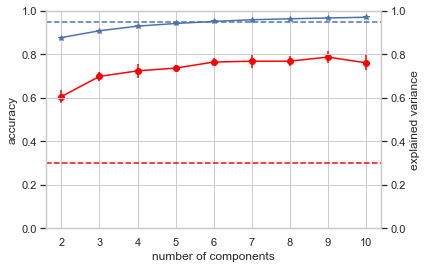} \hspace{0.2cm}
\includegraphics[width=0.47\textwidth]{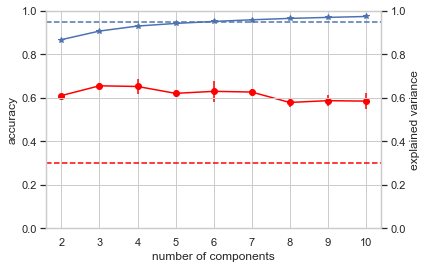}
\end{subfigure}
\vspace{0.1cm}
\caption{Comparison of variation of accuracy (red) with PCA explained variance (blue) as dimension increases for participants 1 (left) and 2 (right). The upper row corresponds to the space of electrodes and the lower row corresponds to the space of sources. The blue dotted line indicates 95\% of explained variance and the red dotted line is chance level. Standard deviations of accuracy (red) are computed after five repetitions of the classifier.}
\label{Fig3-3}
\end{figure} 

The upper row of Fig.\;\ref{Fig3-3} depicts the variation of accuracy with respect to dimension when PCA is applied to the electrode space. A peak is observed in dimensions $3$ and~$4$. By contrast, the lower row in Fig.\;\ref{Fig3-3} compares the variation of accuracy with respect to dimension when PCA is applied to the source space. The graphs suggest that the increase of accuracy (red) is essentially independent of the explained variance by PCA (blue), more visibly in the case of electrodes.

\section{Discussion}
\label{section4}
This study has focused on the design and characterization of a novel
method of data classification by means of topological data analysis.
By contrast to machine learning classifiers, which are based on assessing
relative distance metrics across points of different classes, 
the topological classifier here described is based on exploiting
differences in shape between point clouds from each class.
Operationally, we quantified such 
differences by calculating dissimilarities
between persistence silhouettes~\cite{chazal}. 

The stability theorem for persistence barcodes \cite{chazal, Edelsbrunner2008} guarantees that small variations of a point cloud yield small
variations on 
its persistence descriptors. 
Because of this, if a point cloud is structured into distinct classes ---three in the case of our datasets---
addition of a point to the class where it belongs should be barely noticeable on the resulting topological descriptors, while incorporating the same point
into the other classes is likely to yield more perceivable
shape changes, 
which we quantify and use for 
classification purposes.
We tested this principle with
electro-encephalographic
datasets recorded
from human participants during a study of motivated behaviour
\cite{Cos2021},
aimed at quantifying the influence of three levels of social pressure on the 
specifics of neural dynamics of the human motivational system. 
As for most EEG analyses, the data time series were decomposed into three
frequency bands of interest: $\alpha$, $\beta$ and $\gamma$ (see Methods), 
effectively tripling the analyses, and we also used an unfiltered version of 
the dataset signals for comparison purposes.

\subsubsection*{Choice of dataset}
The choice of dataset responds to several requirements:
\begin{enumerate}
\item The need of having a rich, high-dimensional dataset, which contains data 
clouds with a significant number of elements.
\item As a means to double validate our tests, it was convenient that the dataset
had been previously analyzed by reliable methods, as to offer a clear target of
potential accuracies for the TDA classifier. In our case, we had data from eleven
participants, having performed 1296 trials each, distributed into three classes.
\item Previous machine learning analyses
\cite{Cos2021} had yielded consistent best performances
for the $\gamma$ band across participants, with accuracy ranges between 75-90\%.
\end{enumerate}

\subsubsection*{Dimensionality}
Our study was carried out first with the space of electrodes and subsequently with a space of brain sources obtained from the space of electrodes by means of 
independent component analysis (ICA).
This technique is convenient from a statistical perspective because the dimensions in the source space are statistically independent from each other, and also from a 
neuroscience perspective, as it allows a rough estimation of the brain source 
localization responsible for the data recorded.

We tested the effect of dimensionality by assessing the performance of 
our TDA classifier with an incremental number of dimensions by means of principal component analysis (PCA), hence gradually increasing 
explained variance.
Our PCA analysis ranged up to ten dimensions to 
ensure that we 
covered for over 95\%  of the data variance. This yielded
a remarkable result:
while the degree of explained variance increased
asymptotically over ten dimensions, the TDA classifier's accuracy 
reached a plateau after a certain dimension
---often dimension four in the electrode space.
This strongly suggests that the classifier's operation is more sensitive 
to the latent dimension of the data cloud than to the amount of explained 
variance. This is reinforced by a comparative analysis with the performance of a 1-NN
classifier, which increased its accuracy
with increased explained variance up to dimension ten. 

As an additional validation analysis of our dimensionality tests, we also performed
a recursive feature elimination decomposition (RFE) to identify the signal
dimensions in source space contributing the most to the classification. 
This analysis yielded a similar ceiling effect than PCA, although 
around dimension five.

Our initial prediction was that a projection of the original dataset onto 
an independent component space (or source space) via ICA would yield a finer 
defined cloud and ultimately higher accuracies than those of the original dataset.
This hypothesis is supported by our analysis about the influence of dimensionality 
with the ICA-dataset on the accuracy, which yields a lesser sensitivity when ICA has 
been previously applied to the dataset. Likewise, our validation test with RFE also 
shows that there is a minimal number of necessary sources to obtain an asymptotic
accuracy (typically five).

Ultimately, the number of dimensions that effectively contribute to the classification with
the TDA classifier
match the number of sources that suffice to represent the brain network of 
motivation in the original study 
\cite{Cos2021}.
Moreover, the frequency
band that yielded the best results is the $\gamma$ band, consistently with
previous results.

\subsubsection*{Explained variance versus explained shape}
The accuracy obtained by our TDA-based classifier is comparable with 
those of a nearest neighbout classifer (1-NN) and consistent with previous results
\cite{Cos2021}. However, one of the major differences in performance
is the steady increase of accuracy for the 1-NN classifier with the number of
dimensions used for classification. In other words, the higher the explained variance 
of the signals with which the classifier is trained, the better the accuracy.
Nonetheless, the accuracies for the TDA classifier typically stall after 
a certain dimension consistently across subjects. We believe that the higher sensitivity
of the 1-NN classifier lays in its local nature, as opposed to the global focus
of topological summaries.
In conclusion, we could surmise that TDA provides quantitative methods to assess 
the amount of ``explained shape'' instead of explained variance by a model. Our 
results suggest that the effect of the ambient dimension on the shape of a point 
cloud is of a different nature than its effect on data variance, since shape tends 
to stabilize at some dimension which is intrinsic to the given data.

\section*{Supporting information}

\begin{figure}[H]
\centering
\includegraphics[width=\textwidth]{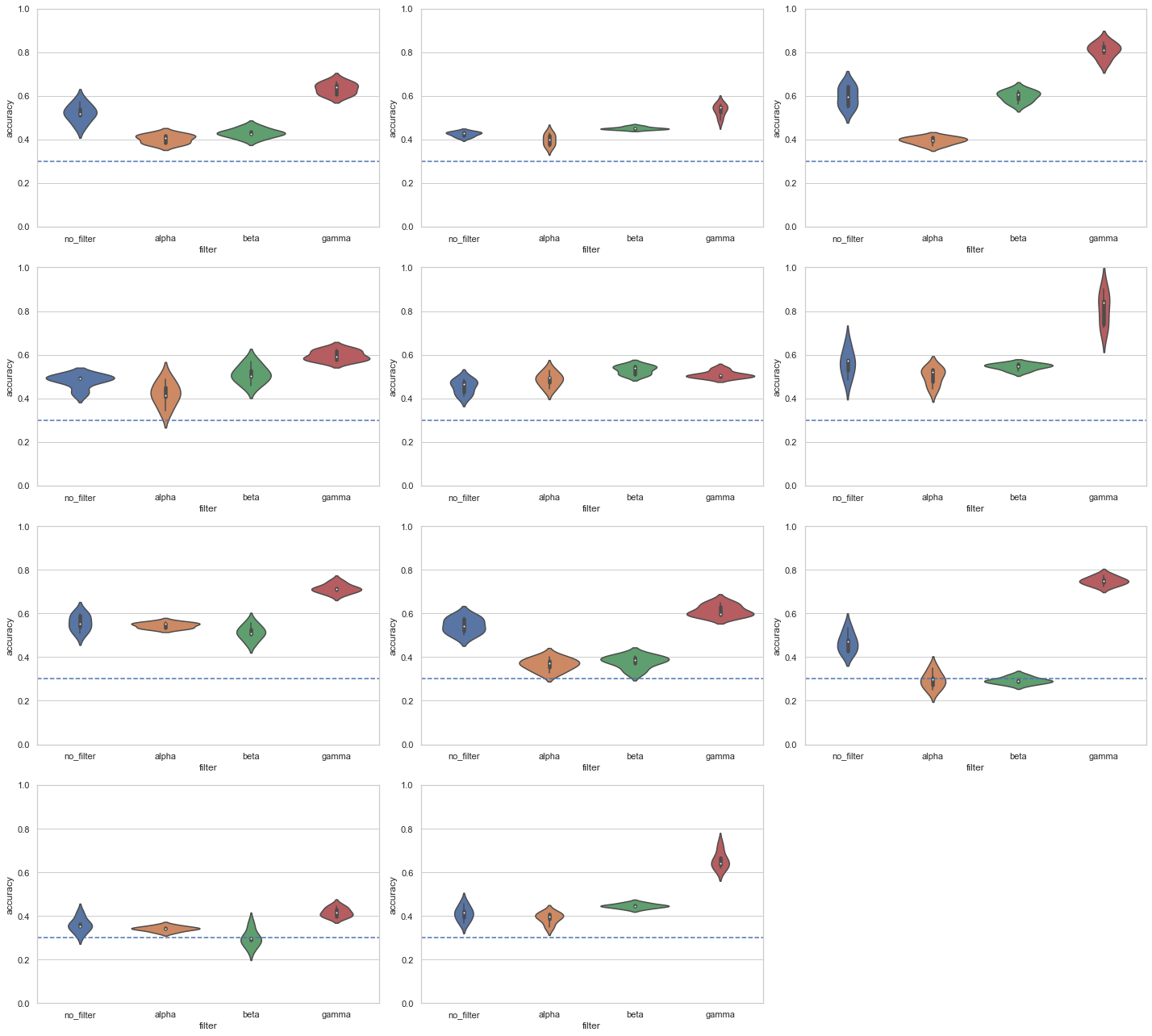}
    \vspace{0.1cm}
    \caption{Accuracies of the topological classifier by frequency band (no filter, alpha, beta, and gamma) on the space of sources for participants 1 to 11 without any dimensionality reduction}
\end{figure}

\begin{figure}[H]
    \centering
    \includegraphics[width=1\textwidth]{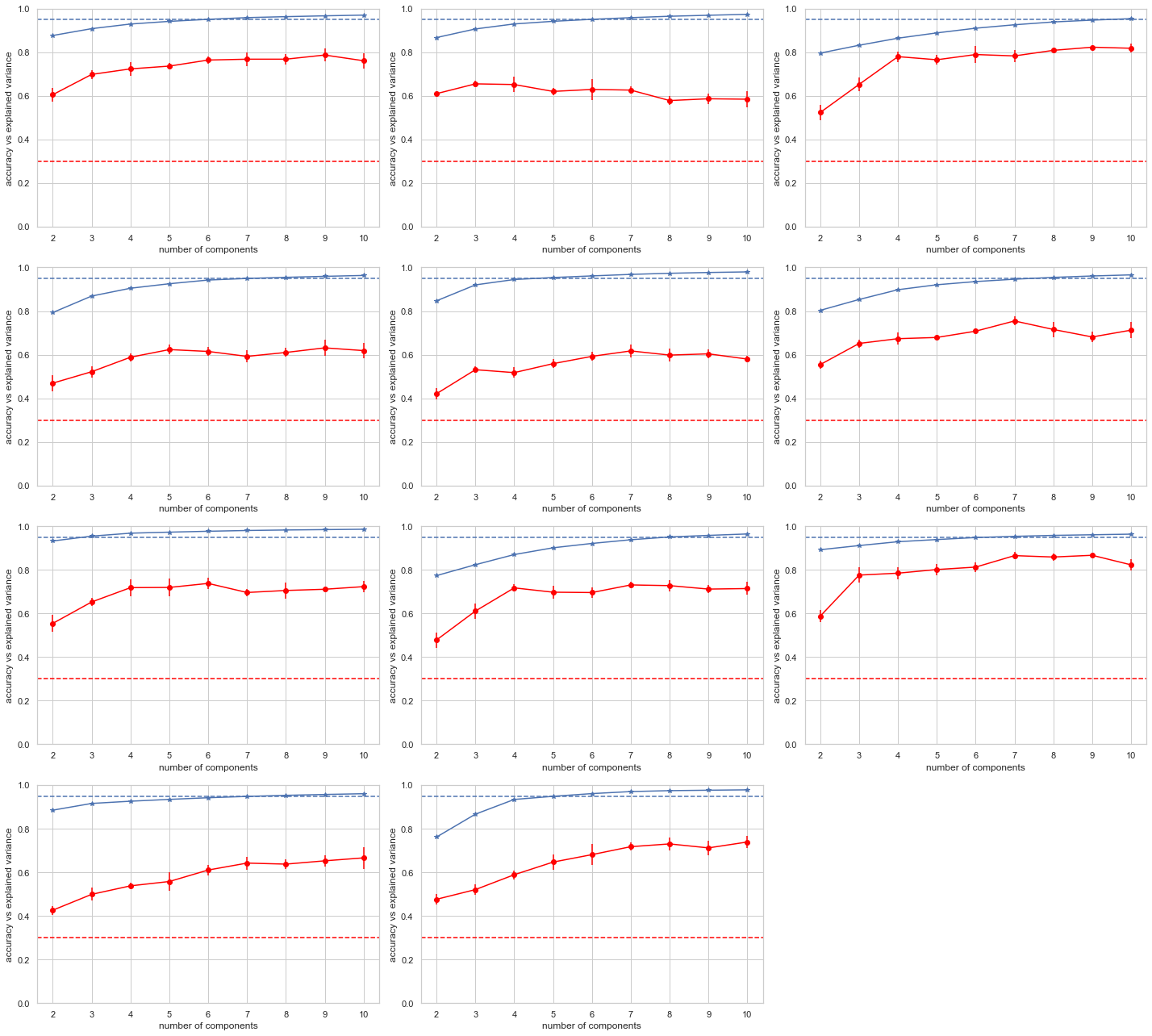}
    \vspace{0.1cm}
    \caption{Comparison of variation of accuracy (red) with \textbf{PCA} explained variance (blue) as dimension increases for all participants on the \textbf{space of sources} within the gamma frequency band. The blue dotted line indicates 95\% of explained variance and the red dotted line is chance level. Standard deviations of accuracy (red) are computed after five repetitions of the classifier.}
\end{figure}

\begin{figure}[H]
    \centering
    \includegraphics[width=1\textwidth]{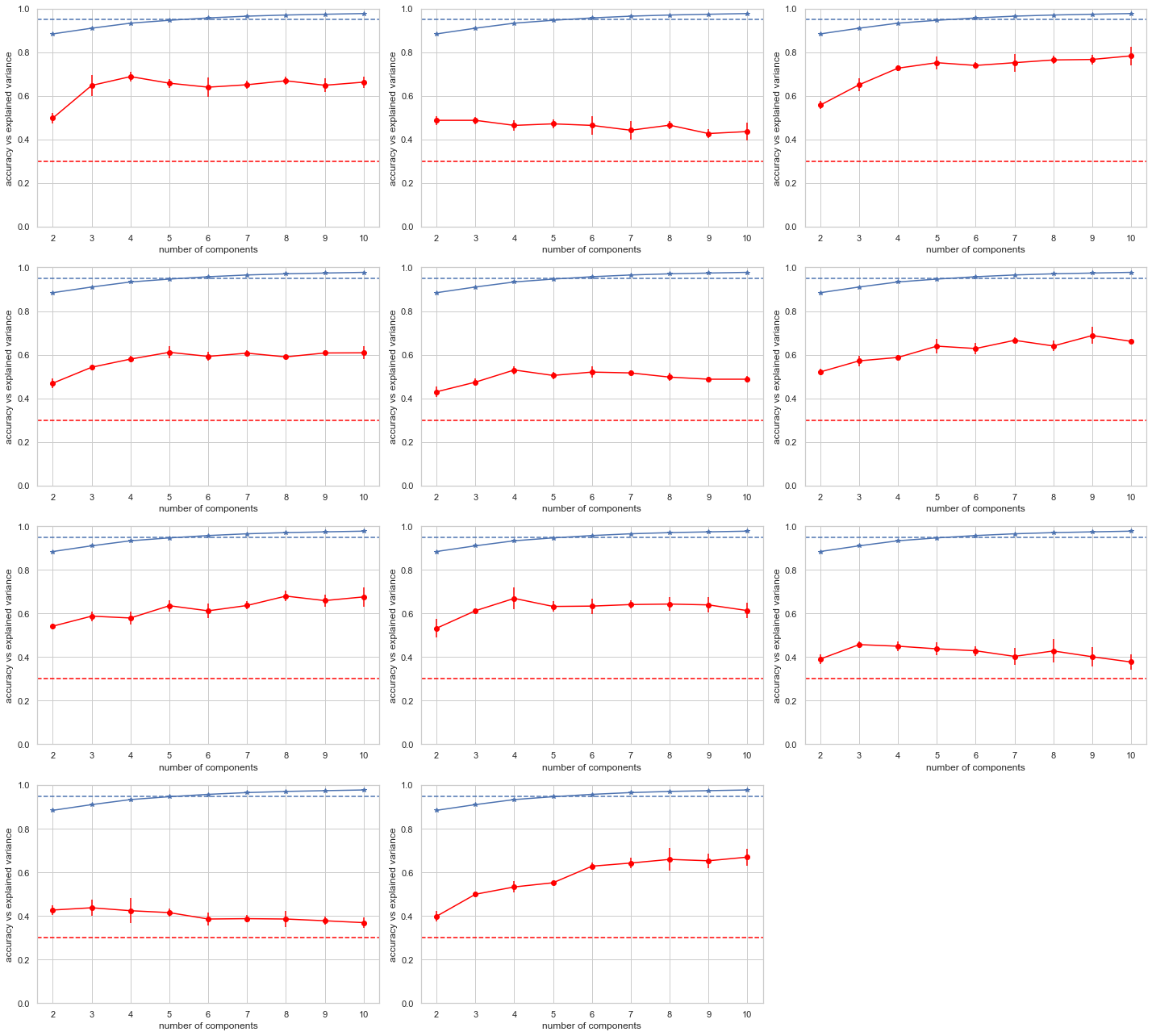}
  \vspace{0.1cm}
    \caption{Comparison of variation of accuracy (red) with \textbf{PCA} explained variance (blue) as dimension increases for all participants on the \textbf{space of electrodes} within the gamma frequency band.}
\end{figure}

\begin{figure}[H]
\centering
\includegraphics[width=1\textwidth]{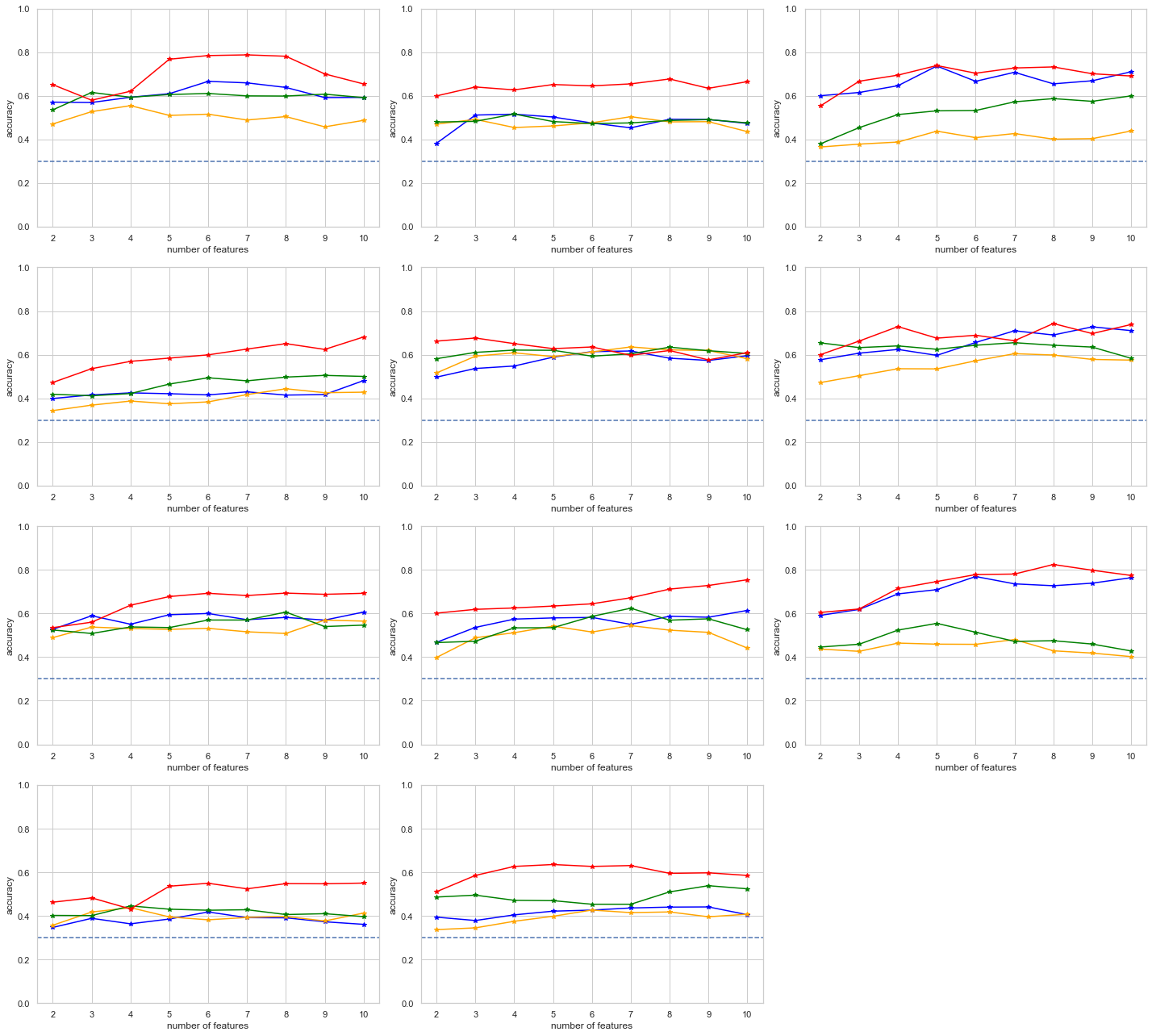}
 \vspace{0.1cm}
    \caption{Comparison of variation of accuracy for each frequency band (blue: no filter; yellow: alpha; green: beta; red: gamma) as the number of sources increases from 2 to 10 by the \textbf{RFE} algorithm, for all participants on the \textbf{space of sources}.}  
\end{figure}

\begin{figure}[H]
\centering
\includegraphics[width=0.9\textwidth]{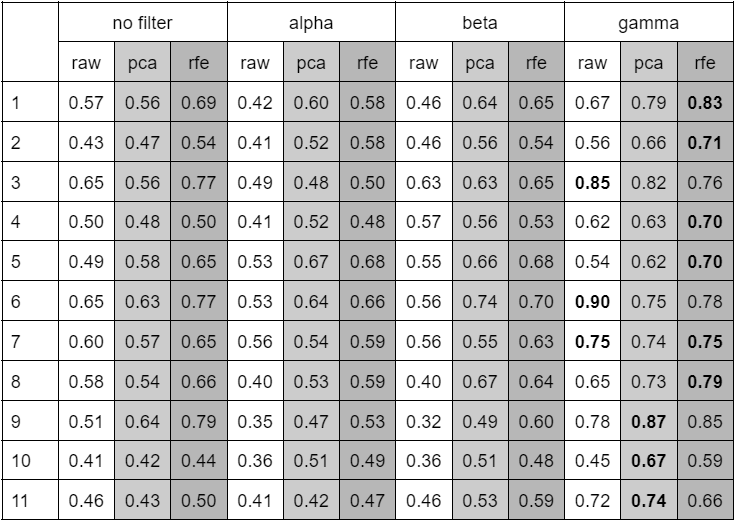}
\vspace{0.3cm}
\caption{Comparison of baseline accuracies (\textit{raw})
of the topological classifier on source space for each frequency band with accuracies obtained after dimensionality reduction with principal component analysis (\textit{pca}) and recursive feature elimination (\textit{rfe}), for participants 1 to 11. The highest accuracies are marked in boldface.} 
\label{Table3-2}
\end{figure}

\begin{table}[H]
\centering
\begin{tabular}{|c|c|c|c|c|c|}
\hline
\multicolumn{3}{|c|}{no filter}                         & \multicolumn{3}{c|}{alpha} \\ \hline

raw & pca & rfe 
& raw & pca & rfe \\ \hline

$0.57 \pm 0.04$ & $0.56 \pm 0.06$ 
& $0.69 \pm 0.04$ & $0.42 \pm 0.02$ 
& $0.60 \pm 0.01$ & $0.58 \pm 0.04$  \\ \hline

$0.43 \pm 0.01$ & $0.47 \pm 0.02$ 
& $0.54 \pm 0.05$ & $0.43 \pm 0.03$ 
& $0.52 \pm 0.03$ & $0.58 \pm 0.04$  \\ \hline

$0.65 \pm 0.05$ & $0.56 \pm 0.03$ 
& $0.77 \pm 0.06$ & $0.41 \pm 0.02$ 
& $0.48 \pm 0.02$ & $0.50 \pm 0.04$  \\ \hline

$0.50 \pm 0.03$ & $0.48 \pm 0.04$ 
& $0.50 \pm 0.03$ & $0.49 \pm 0.05$ 
& $0.52 \pm 0.02$ & $0.48 \pm 0.04$  \\ \hline

$0.49 \pm 0.03$ & $0.56 \pm 0.03$ 
& $0.65 \pm 0.04$ & $0.53 \pm 0.03$ 
& $0.67 \pm 0.03$ & $0.68 \pm 0.04$  \\ \hline

$0.65 \pm 0.06$ & $0.63 \pm 0.05$ 
& $0.77 \pm 0.06$ & $0.53 \pm 0.04$ 
& $0.64 \pm 0.02$ & $0.66 \pm 0.05$  \\ \hline

$0.60 \pm 0.04$ & $0.57 \pm 0.06$ 
& $0.65 \pm 0.04$ & $0.56 \pm 0.01$ 
& $0.54 \pm 0.02$ & $0.59 \pm 0.03$  \\ \hline

$0.58 \pm 0.03$ & $0.54 \pm 0.03$ 
& $0.66 \pm 0.05$ & $0.40 \pm 0.03$ 
& $0.53 \pm 0.02$ & $0.59 \pm 0.06$  \\ \hline

$0.54 \pm 0.05$ & $0.64 \pm 0.04$ 
& $0.79 \pm 0.06$ & $0.35 \pm 0.04$ 
& $0.47 \pm 0.03$ & $0.53 \pm 0.03$  \\ \hline

$0.41 \pm 0.03$ & $0.42 \pm 0.06$ 
& $0.44 \pm 0.03$ & $0.36 \pm 0.01$ 
& $0.51 \pm 0.03$ & $0.49 \pm 0.03$  \\ \hline

$0.46 \pm 0.03$ & $0.43 \pm 0.01$ 
& $0.50 \pm 0.04$ & $0.41 \pm 0.03$ 
& $0.42 \pm 0.02$ & $0.47 \pm 0.04$  \\ \hline
\end{tabular}
\vspace{0.3cm}
\caption{Comparison of baseline accuracies (\textit{raw})
of the topological classifier on source space for each frequency filter with accuracies obtained after dimensionality reduction with principal component analysis (\textit{pca}) and recursive feature elimination (\textit{rfe}), with no frequency filter and in the alpha frequency band, for patients 1 to 11 with standard deviations after five repetitions.}
\label{nofilter-alpha}
\end{table}

\begin{table}[H]
\centering
\begin{tabular}{|c|c|c|c|c|c|}
\hline
\multicolumn{3}{|c|}{beta}                              & \multicolumn{3}{c|}{gamma}                             \\ \hline

raw & pca & rfe 
& 
raw & pca & rfe \\ \hline

$0.46 \pm 0.02$& $0.64 \pm 0.03$ 
& $0.65 \pm 0.03$ & $0.67 \pm 0.03$ 
& $0.79 \pm 0.03$ & $0.83 \pm 0.08$ \\ \hline

$0.46 \pm 0.01$ & $0.56 \pm 0.03$ 
& $0.54 \pm 0.03$ & $0.56 \pm 0.03$ 
& $0.66 \pm 0.01$ & $0.71 \pm 0.04$ \\ \hline

$0.63 \pm 0.03$ & $0.63 \pm 0.02$ 
& $0.65 \pm 0.07$ & $0.85 \pm 0.03$ 
& $0.82 \pm 0.01$ & $0.76 \pm 0.06$ \\ \hline

$0.57 \pm 0.04$ & $0.56 \pm 0.02$ 
& $0.53 \pm 0.04$ & $0.62 \pm 0.02$ 
& $0.63 \pm 0.04$ & $0.70 \pm 0.07$ \\ \hline

$0.55 \pm 0.02$ & $0.66 \pm 0.04$ 
& $0.68 \pm 0.02$ & $0.54 \pm 0.02$ 
& $0.62 \pm 0.03$ & $0.70 \pm 0.04$ \\ \hline

$0.56 \pm 0.01$ & $0.74 \pm 0.02$ 
& $0.70 \pm 0.03$ & $0.90 \pm 0.08$ 
& $0.75 \pm 0.02$ & $0.78 \pm 0.05$ \\ \hline

$0.56 \pm 0.03$ & $0.55 \pm 0.01$ 
& $0.63 \pm 0.04$ & $0.75 \pm 0.02$ 
& $0.74 \pm 0.02$ & $0.75 \pm 0.06$ \\ \hline

$0.40 \pm 0.03$ & $0.67 \pm 0.02$ 
& $0.64 \pm 0.05$ & $0.65 \pm 0.03$ 
& $0.73 \pm 0.02$ & $0.79 \pm 0.06$ \\ \hline

$0.32 \pm 0.01$ & $0.49 \pm 0.03$ 
& $0.60 \pm 0.05$ & $0.78 \pm 0.02$ 
& $0.87 \pm 0.01$ & $0.85 \pm 0.08$ \\ \hline

$0.36 \pm 0.04$ & $0.51 \pm 0.03$ 
& $0.48 \pm 0.03$ & $0.45 \pm 0.02$ 
& $0.67 \pm 0.05$ & $0.59 \pm 0.05$ \\ \hline

$0.46 \pm 0.01$ & $0.53 \pm 0.02$ 
& $0.59 \pm 0.04$ & $0.72 \pm 0.04$ 
& $0.74 \pm 0.03$ & $0.66 \pm 0.04$ \\ \hline
\end{tabular}
\vspace{0.3cm}
\caption{Comparison of baseline accuracies (\textit{raw})
of the topological classifier on source space for each frequency filter with accuracies obtained after dimensionality reduction with principal component analysis (\textit{pca}) and recursive feature elimination (\textit{rfe}), in the beta and gamma frequency bands, for patients 1 to 11 with standard deviations after five repetitions.}
\label{beta-gamma}
\end{table}

\end{document}